\documentclass{bmvc2k}
\usepackage{enumitem}

\newcommand{\nmpar}[1]{\hfill\break\noindent\textbf{#1}}

\title{Image Retrieval with Mixed Initiative and Multimodal Feedback}

\addauthor{Nils Murrugarra-Llerena}{nineil@cs.pitt.edu}{1}
\addauthor{Adriana Kovashka}{kovashka@cs.pitt.edu}{1}

\addinstitution{
 Department of Computer Science\\
 University of Pittsburgh\\
 Pittsburgh, PA, USA
}

\runninghead{Murrugarra, Kovashka}{Image Retrieval with Mixed Initiative}

\begin{document}

\maketitle

\begin{abstract}
%\nm{In order to engage online buyers, online shopping websites allow interactive search.} 
How would you search for a unique, fashionable shoe that a friend wore and you want to buy, but you didn't take a picture?
Existing approaches propose interactive image search as a promising venue. However, they either entrust the user with taking the initiative to provide informative feedback, or give all control to the system which determines informative questions to ask. 
Instead, we propose a \emph{mixed-initiative} framework where both the user and system can be active participants, depending on whose initiative will be more beneficial for obtaining high-quality search results. 
We develop a reinforcement learning approach which dynamically decides which of three interaction opportunities to give to the user: drawing a sketch, providing free-form attribute feedback, or answering attribute-based questions. 
By allowing these three options, our system optimizes both the informativeness and
exploration capabilities allowing faster image retrieval. % of the search interaction, usually early on in the search.
% We verify the usefulness of our approach on three dataset and extensive experimental settings.
We outperform three baselines on three datasets and extensive experimental settings. %\nm{We improve upon prior approaches by up 3\% in AUC percentile rank.}
\end{abstract}

\vspace{-0.3cm}
\section{Introduction}
\label{sec:intro}

%1) need to truly capture user's state
%2) some prior work but it's limited
%3) what we do: allow user to explore/exploit, use multiple modalities, thus capture their idea better, in a more active way
%4) preview of results

\begin{figure}[t]
    \centering
    \includegraphics[width=1\textwidth]{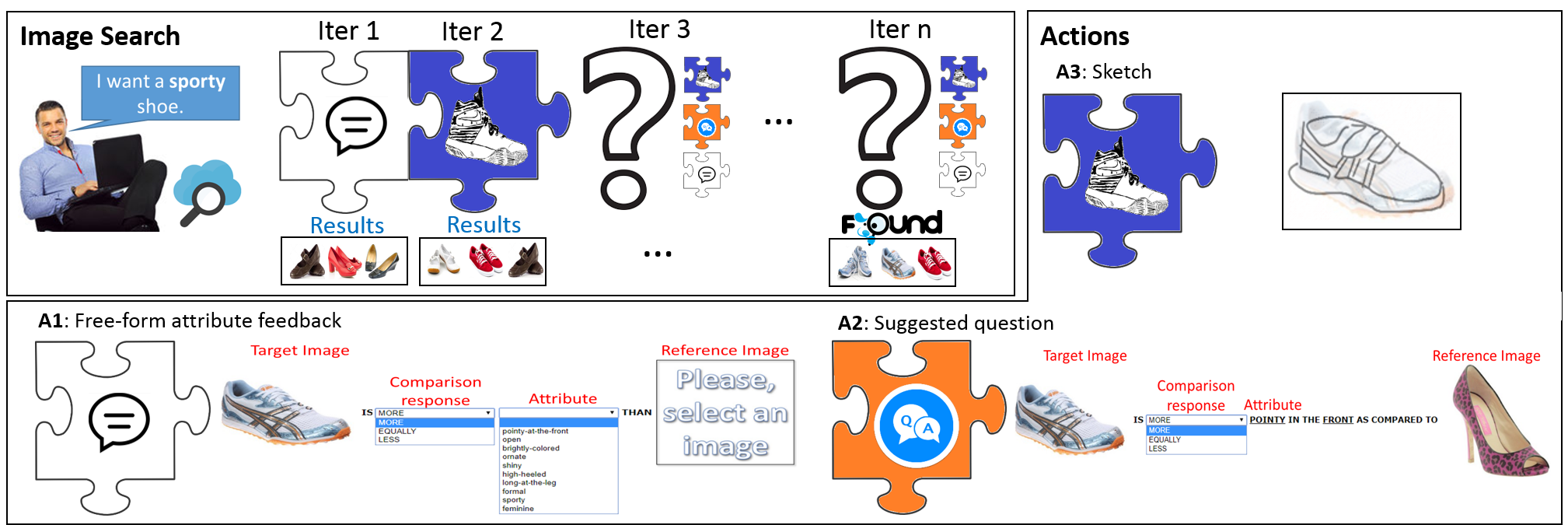}
    \vspace{-0.5cm}
    \caption{We learn how to intelligently combine different forms of user feedback for interactive image search, and find the user's desired content in fewer iterations. The \textit{image search} section depicts our search agent that predicts an appropriate action at a certain iteration. %These decisions will help to find more images in less number of iterations. 
    % In this example, the first two predicted actions are free-form attribute feedback and sketch. 
    Then, the \textit{actions} section presents the three possible interactions (actions) of our agent.
    }
    \label{fig:concept}
    \vspace{-0.5cm}
\end{figure}
\vspace{-0.3cm}

Computer vision apps serve a variety of user needs: for example, they can automatically count calories \cite{Meyers_2015_ICCV}, summarize vacation footage \cite{Yao_2016_CVPR},  ``paint''  \cite{Gatys_2016_CVPR}, or
help users find shoes they want to buy \cite{kovashka2015whittlesearch} via image search. While for calorie-counting or machine-painting
the interaction between the user and the machine is limited to submitting a photograph, for image search the user
needs to communicate with the system in a more fine-grained and unrestricted fashion, since success is defined by
whether the system successfully ``guessed'' what the user wanted to find. A person can look for online shopping options on products they saw in a store, or even try to find a criminal they saw in an online database. 
The user's mental concept of what they wish to
retrieve can be arbitrarily subtle hence difficult to capture, and in order to ensure that the system's model of the user's
search concept is accurate, the user needs to be able to ``explain'' to the system how it should adjust its predictions.

Prior work has tackled this challenge in a number of ways. 
Some work has used semantic visual attributes (like ``shiny'' or ``chubby'') \cite{Lampert09,Farhadi09,parikh2011relative,kovashka2015whittlesearch} to allow the user to give precise language-based guidance to the system. Attributes provide an excellent
channel for communication because humans naturally explain the world to each other with adjective-driven descriptions. Attributes have been shown
promising as a tool for image search \cite{kumar2011describable,siddiquie2011image,kovashka2015whittlesearch,Yu_2015_ICCV,huang2015cross,Prabhu_2015_ICCV}. 
For example,
\cite{kovashka2015whittlesearch} show how a user can perform
rich relevance feedback by specifying how the attributes of a results image should change to better match the user's
target image. For example, the user might say ``Show me people with longer hair than this one.'' 
Another approach has been to engage the user in question-answering with questions that the system estimated are most useful \cite{kovashka2013pivots,ferecatu2009statistical}. 
Thus, in prior work, the initiative for what guidance to give to the system has been taken by either the user \cite{kovashka2012whittlesearch,kovashka2015whittlesearch,kumar2011describable,siddiquie2011image,Yu_2015_ICCV} or system \cite{ferecatu2009statistical,tong2001support,kovashka2013pivots} \emph{but not both.}
Another approach has been to allow the user to provide visual cues for what they are looking for, e.g. by drawing a sketch \cite{eitz2011sketch,yu2016sketch,yu2017sketch,sangkloy2016sketchy}. The system can then retrieve visually similar results. 
Thus, the user can use either language or visuals to search, but it is not clear which modality is more informative. 

%we assume user doesn't have exact image...

In our work, we propose a framework where \textbf{either} the user \textbf{or} system can drive the interaction, and the input modality can be \textbf{either} textual \textbf{or} visual, depending on what seems most beneficial at any point in time. 
For example, the user can kick off the search using a sketch, then refine the results by explaining how the top retrieved images at a certain iteration differ from her mental model. 
%\nm{from top result images at a certain iteration.} %\ak{not clear what these few images are, are they reference/feedback images?} 
Then the system might ask attribute-based questions, and give control back to the user when it runs out of informative questions to ask, so the user can provide some more free-form attribute feedback of their choosing.
Since it is the system that must rank the results, we propose to leave the choice of what is most informative to the system. 
In other words, the system can decide to let the user lead and \emph{explore}, if it cannot \emph{exploit} any relevant information in a certain iteration. %the basic information it already has using actively chosen queries. 
The system can encourage the user to provide multimodal feedback, i.e. textual \textbf{or} visual feedback. 
To make all these decisions, we train a reinforcement learning agent (see Fig.~\ref{fig:concept}).

In particular, the options that the reinforcement learning chooses between are: (1) sketch feedback, (2) free-form attribute feedback, or (3) system-chosen attribute questions.
At each iteration, the system adaptively chooses one of these interactions and asks the user to provide the corresponding type of feedback (e.g. it asks the user to choose an image and attribute to comment on).
%, or it selects the image and attribute and asks the user a question about those).
Briefly, our method works as follows. 
Our agent receives a state composed of top result images, proxies for the target image, 
%\footnote{which encapsulates information about the target image}, 
and history of taken actions, when available. It interacts with the environment trying different actions. Over time, it learns to pick the most meaningful action, given a certain state. We guide our agent with information about whether the target image is among our top results.

Note that while it is the system that decides what \emph{type} of search interaction is most useful, both the user and system are \emph{active} participants in the search. When the system gives control back to the user, the user can freely choose what language-based feedback to provide or what imagery to sketch.
This is in contrast to prior work involving human-machine search interactions where only one party, either the user \cite{kovashka2012whittlesearch,kovashka2015whittlesearch,kumar2011describable,siddiquie2011image,Yu_2015_ICCV} or system \cite{kovashka2013pivots}, drives the search. 
In contrast, in \emph{human-to-human} interactions, the participants in a conversation trade off
control: usually all participants at least have the possibility of both asking and answering questions.
To combine the information-theoretic benefits of \cite{kovashka2013pivots} and the explorative nature of interaction of \cite{kovashka2015whittlesearch,eitz2011sketch}, we propose
a framework that allows the machine and human to alternate, depending on who can initiate more informative feedback.

\vspace{-0.3cm}
\section{Related Work}
\label{sec:related}
\vspace{-0.1cm}

\emph{Attribute-based search.}
Prior work has explored the value of the fine-grained detail that attribute descriptions provide, by using attributes to initiate a search \cite{siddiquie2011image,vaquero2009attribute} or provide iterative feedback on the results of a search system \cite{kovashka2015whittlesearch,kovashka2013pivots,modi2017confidence}.
\cite{kovashka2012whittlesearch} browses the current search results, and can then provide a feedback statement of the form ``The image I am looking for is more/less [attribute] than [this image in the results].'' The choice of an attribute on which to comment is left to the user. This is helpful %, which has advantages 
if the user is perceptive, or there are images which obviously differ from the user's desired content for particular attributes. On the other hand, browsing a set of images and choosing attributes is time-consuming for the user, as we find in experiments. \cite{kovashka2013pivots} shows that given a limited budget of interactions that the user is willing to perform, more accurate search results can be achieved if the system asks the user questions of the form ``Is the image you are looking for more/less/equally [attribute] than [this image]?'' The chosen questions are those with high information gain. 
%Note that \cite{kovashka2013pivots} can actively choose the questions to ask in real time, as it narrows down the search for potential images to those that split an attribute search space in half.  
The disadvantage of \cite{kovashka2013pivots} is that it limits the ability of the user to browse and explore the dataset space.

\emph{Sketch-based search.}
While attribute-based feedback is appropriate when the user can concisely describe what content they wish to find using words, some searches involve concepts which are purely visual. In our setting, we assume the user does not have a photograph of what they wish to find, so cannot directly do similarity-based search with a query image. However, the user does have a clear visual idea of what content they wish to find. Sketch-based search approaches allow the user to convey this visual idea to the system, via a sketch or drawing, which provides a complementary way of communication. The system can then extract features from this sketch and compare to the features of the images in a database \cite{eitz2011sketch,shrivastava2011data,yu2016sketch,sangkloy2016sketchy,yu2017sketch}.
We use a similar approach, but also propose to convert the sketch to an image using generative models. 
Other authors use generative learning to find a representation appropriate for cross-domain (sketch-to-image \cite{pang2017cross, song2017retrieval, song2017deep} or text-to-image \cite{song2017retrieval}) search. 
%\cite{song2017deep} focus on fine-grained sketch-to-image retrieval using an attention mechanism to avoid misalignment.
% \cite{pang2017cross} use generative learning to find a representation appropriate for cross-domain (sketch-to-image) search. Also, \cite{song2017retrieval} find an intermediate representation for text-to-image and sketch-to-image search. 
%However, none of these approaches do not explicitly convert the sketch to an image during retrieval.
%, but do not explicitly convert the sketch to an image during retrieval.
We use sketch-based retrieval in a larger reinforcement learning framework that chooses which search interaction to propose (sketch, attribute-based feedback, or question-answering). 
Note that our focus is \emph{not} in how we perform sketch-based retrieval, but rather \emph{how to decide when} to request a sketch.

\emph{Interactive search.}
Rather than ask the user to issue a query and return a single set of results, we engage the user in providing interactive relevance feedback and show results after each round. 
This is a popular idea \cite{rui1998relevance,zhou2003relevance,cox1996pichunter,fogarty2008cueflik,ferecatu2009statistical} whose key benefit is that incorrect predictions by the system can be corrected.
% geman - give examples of questions
We also adopt interactive search, but combine the advantages of free-form feedback and exploration with the information-theoretic benefits of actively querying for feedback \cite{ferecatu2009statistical}, via reinforcement learning.

\emph{Active learning.}
In order to minimize the cost of data labeling, active learning approaches estimate the potential benefit of labeling any particular image, using cues such as entropy, uncertainty reduction, and model disagreement
\cite{tong2001support,seung1992query,guo2007optimistic,vijayanarasimhan2014large,kading2015active}.
\cite{wolfman2001mixed,cakmak2011mixed,suh2016label,ebert2012ralf} have explored mixed initiative between user and system as well as reinforcement learning, for improving active learning at training time, in contexts other than image search.
In contrast, we use reinforcement learning to select interactions at test time (during online search).

\emph{Reinforcement learning}
 \cite{kaelbling1996reinforcement,mnih2015human,van2016deep} has recently gained popularity for a variety of computer vision tasks, e.g. object  \cite{caicedo2015active,mathe2016reinforcement} and action detection \cite{yeung2016end}.
The most related work to ours is \cite{yin2005integrating} which also uses reinforcement learning to choose the type of feedback method for requesting feedback from the user. This approach considers query vector modification, feature relevance estimation, and Bayesian inference, as three possible feedback mechanisms. Neither of these allows the user to \emph{comparatively} describe how the results should change (via attributes); instead, each image property is defined as desirable/undesirable.  \cite{kovashka2015whittlesearch} show such binary feedback is inferior to comparative attribute feedback.  
Further, unlike \cite{yin2005integrating}, we consider both visual and textual feedback among the mechanisms presented to our users.

%\emph{Generative models.}
%Generative adversarial networks \cite{goodfellow2014generative} have allowed many applications converting from one modality to another \cite{Isola_2017_CVPR,zhu2017unpaired,choi2017stargan}. We use them to extrapolate what a sketch might look like, so we can retrieve images similar to the photo version of the sketch. 

\vspace{-0.3cm}
\section{Approach}
\label{sec:approach}
\vspace{-0.1cm}

We develop an approach for interactive image retrieval, where the user can provide guidance to the system via two text-based and one sketch-based modalities, described below. The search scenario we envision is the following: The user has a clear idea of the exact target image they wish to find, but does not have that image in hand. Our system's goal is to determine which type of interaction to suggest to the user at any point in time. %\nm{for an accurate search.}

\vspace{-0.3cm}
\subsection{Search setup and interactions} 
\label{sec:setup}
\vspace{-0.1cm}

\paragraph{Interactions.}
The user can initiate a search with random images from the database, or ones that match a simple keyword query. Then the user can perform a combination of the following three types of feedback. First, the user can browse the returned images, and relate them to her desired target via attribute comparisons, e.g. ``The person I am looking for is \emph{younger} than this person,'' where ``this person'' is an image chosen from the returned results. Second, the system can ask the user a question, e.g. ``Is the person you are looking for more or less chubby than this person?'' Third, the user can draw a sketch to visually convey to the system their desired content. These search interactions are based on prior work \cite{kovashka2015whittlesearch,kovashka2012whittlesearch,kovashka2013pivots,eitz2011sketch,sangkloy2016sketchy,yu2016sketch}, and we learn how to combine them.
%\emph{Our goal is to develop an approach for choosing which type of interaction should be used at any point in time.}

\begin{figure}[h!]
    \centering
    \includegraphics[width=0.9\textwidth]{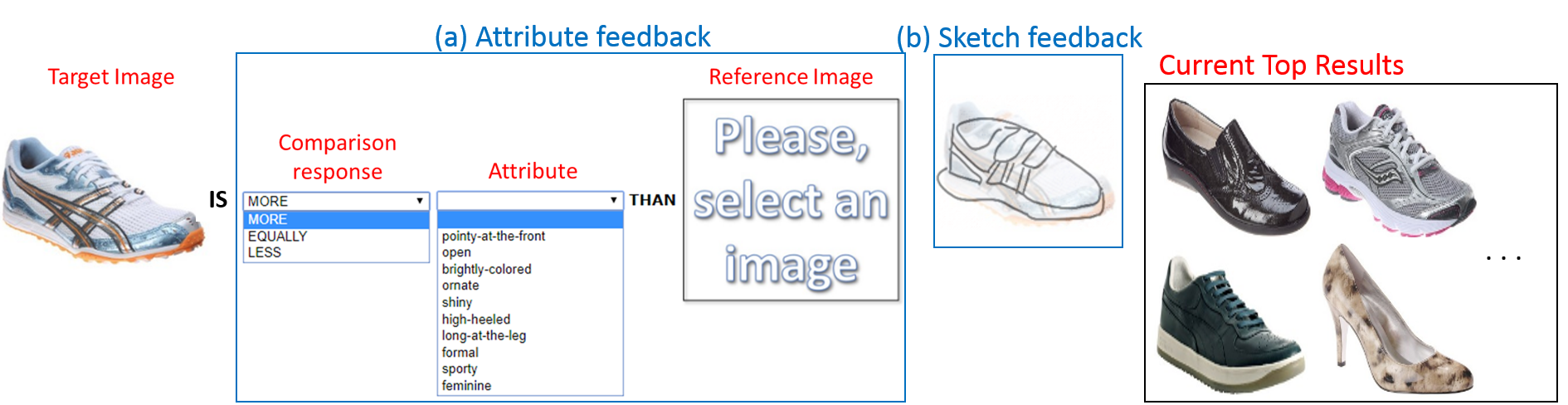}
    \vspace{-0.1cm}
    \caption{Image retrieval system setup. The system's goal is to find the target image. Users refine the image retrieval using an (attribute, reference image, comparison response) triplet or a sketch. User interactions are used to update the current top image results. %\ak{Leave a bit more space after "IS" before the first image.}
    } 
    \label{fig:example_interactive_search}
    \vspace{-0.6cm}
\end{figure}

\vspace{-0.4cm}
\paragraph{System interface.} 
Our system is illustrated in Figure \ref{fig:example_interactive_search}, and it has three components: i) a target image, ii) user feedback using attributes or a sketch, and iii) current top images. User feedback is received in each iteration, and updates the top images. 
%This feedback has two forms: an attribute-based feedback, or a sketch one. The attribute feedback is composed of a comparison question that involves the target image, an attribute,  and a reference image. The attribute feedback can be provided in two different ways, one driven by the system \cite{kovashka2013pivots} which suggests an (attribute, reference image) pair, and one driven by the user \cite{kovashka2015whittlesearch} where the user selects the image-attribute pair. On the other hand, the sketch feedback only consists of a drawing of the target image.
%We briefly introduce how our image retrieval system works, and we provide the Markov Decision Process (MDP) for our reinforcement learning agent.

\vspace{-0.4cm}
\paragraph{Relevance models.} After one of the three interactions is used and feedback from the user is received, the system must rank all database images by estimating their relevance using the feedback the user provided. For free-form attribute feedback and suggested question interactions, following \cite{kovashka2015whittlesearch}, the relevance of a database image is proportional to the likelihood that it satisfies each attribute constraint, e.g. it is more shiny than a reference image.
%: [add equation] 
For sketch interaction, 
we ``convert'' the sketch to a photograph (i.e. we add color) using a conditional GAN \cite{Isola_2017_CVPR}. 
An alternative is to directly learn a space whether sketches and images are aligned, and perform retrieval in this space; we show an experiment using this approach as well.
Then, CNN features are extracted and we train a one-class SVM \cite{Scholkopf2001} whose output probabilities for each image are used to rank the images.
%: [add equation] 
The final relevance of an image is a product (multiplication) of all attribute-based and sketch-based relevance estimates.
%: [add equation]

\vspace{-0.3cm}
\subsection{Reinforcement learning representation}
\label{sec:repr}
\vspace{-0.1cm}

We formulate the selection over search interactions as a Markov Decision Process composed of actions, states, and rewards, defined below.

\vspace{-0.4cm}
\nmpar{Actions.} 
We train a reinforcement learning algorithm to select one of three interactions for a given iteration. In order to train it, we require user selections of image-attribute pairs (the free-form feedback proposed in \cite{kovashka2012whittlesearch}), responses to attribute-based questions proposed in \cite{kovashka2013pivots} (the more/less/equally value of a comparison between the target and reference image along a certain attribute dimension), and sketches (used for search in \cite{yu2016sketch,eitz2011sketch,sangkloy2016sketchy}). User selections are simulated by selecting an (image, attribute) pair that reduces the part of the multi-attribute space that needs to be searched in order to find the target image. In particular, our simulated users are given a subset of the attribute vocabulary\footnote{Since our simulated users receive system-level information as described next, allowing them to use the full vocabulary results in unrealistic alignment between the user's mental model and the system's predictions.}, and a set of reference images. They are also given information about how many images in the database satisfy a given image-attribute constraint, e.g. how many images are ``less chubby than [this person],'' according to the system's model of ``chubbiness.'' 
The simulated user then chooses the image-attribute pair that results in the smallest number of images satisfying the constraint. This simplifies search as only a few images remain relevant after each feedback constraint is given.

%Our actions are categorized as attribute free-form feedback, system-suggested questions about attributes, and sketch feedback. 
%Hence, each image has an associated probability, and can be easily integrated with \textit{user drives} and \textit{system drives}.
% \nm{(need to explain normalization procedures)}. 
%The (image, attribute) pair is selected among a subset of attributes and top images. 
%we can perform a human in the loop solution, however, this requires user annotations and it is costly. 

In terms of question responses, 
we also simulate users' feedback, similarly to \cite{kovashka2013pivots}, by adding Gaussian noise to the attribute model predictions, and choosing the more/less/equally value based on the difference in the attribute values predicted for the target image and the reference image which the system chose.
The original method of \cite{kovashka2013pivots} requires entropy computation, which is computationally expensive if it needs to be repeated many times, as we require for reinforcement learning. 
Hence, we use an ablation presented in \cite{kovashka2013pivots} which performs similarly but is much faster. It uses the per-attribute binary search trees of \cite{kovashka2013pivots} but alternates between attribute pivots in a round-robin fashion. 

Sketches are simulated using edge maps \cite{Xie_2015_ICCV} generated from the target image, similarly to \cite{Isola_2017_CVPR}. We also show experiments using real human-drawn sketches. We then convert them to photographs using a GAN \cite{Isola_2017_CVPR}, and rank database images by their similarity to the photo, using the probabilities from a one-class SVM \cite{Scholkopf2001}.

\vspace{-0.4cm}
\nmpar{State.}
Let $h_{+prox}$ and $h_{-prox}$ be positive and negative proxy sets for the target image, defined as the five neighbors closest to the target (excluding the target itself), and five neighbors furthest from the target. 
We represent our state as ($h_{top\_ims}$, $h_{+prox}$, $h_{-prox}$, $h_{actions}$), where $h_{top\_ims}$ is the history of top images (i.e. those ranked at the top in previous iterations), and $h_{actions}$ are the actions taken in previous iterations.
Images are represented by features extracted from AlexNet \cite{NIPS2012_4824}, and actions by a 3-dimensional binary vector, where all values are zero, except the one corresponding to the taken action. We use a history size of 3.

\vspace{-0.4cm}
\nmpar{Rewards.} We would like that in each iteration, our top images become more and more similar to the target image (which is unknown to the system). We can measure this using two cues: distance to positive proxy images, and distance to negative proxy images. We encourage a decrement of the first distance, and an increment of the later distance. We do this using a reward function $R(s, s')$ which is evaluated when an action is performed and causes a transition from state $s$ to state $s'$. Each state has associated top images ($top\_ims$) and proxies ($+prox$ and $-prox$). Then the function $R$ is defined as:

\vspace{-0.5cm}
\begin{equation}
\footnotesize
    R(s, s') = sign[d(top\_ims, +prox) - d(top\_ims', +prox)] + sign[d(top\_ims', -prox) - d(top\_ims, -prox)]
    \label{eq:reward}
%\vspace{-0.2cm}
\end{equation}

In other words, we want the distance of the top images to positive proxies to decrease, and distance to negative proxies to increase.  
We calculate the Euclidean distance $d$ between (1) the average features of all top images and (2) the average features of the positive/negative proxy images. We also want to encourage that the sketch action is used only once%\nm{, because it is common to use only one sketch query}
. Hence, we assign a penalty of $-1$ if the sketch interaction is requested more than once.

\vspace{-0.3cm}
\subsection{Learning}
\vspace{-0.1cm}

\begin{figure}[t]
    \centering
    \includegraphics[width=1.0\textwidth]{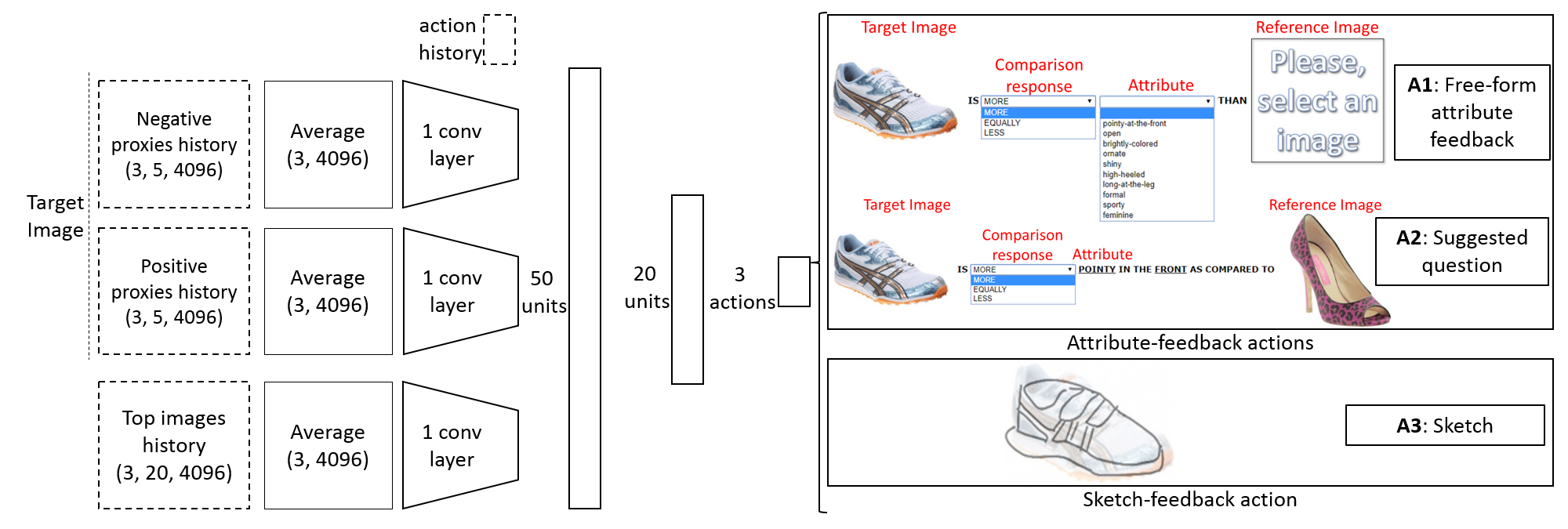}
    \vspace{-0.8cm}
    \caption{Architecture of our proposed Q-network. It receives histories of top-ranked images, positive and negative proxy images, and taken actions.
     %The former three inputs are processed by 1 convolutional layers. Then, all these three branches are concatenated with action history, and fed to 3 fully connected layers. Hence, our Q-network 
     It predicts the best action given a specific state. Inputs are denoted with dotted lines. Please see text for further explanation. %\ak{Make title for A1 be free-form ATTRIBUTE feedback, for consistency with Fig. 1. For consistency with the retrieval experiments, here and in all figs, can you fade (e.g. make partially transparent) the GAN-generated image from the sketch method? Also please make the right-hand side smaller so it's as tall as the left-hand side, to reduce use of whitespace.}
     } 
    \label{fig:q_network}
\vspace{-0.3cm}
\end{figure}

The goal of our agent is to update the search results by selecting actions. There are many possible states, so using a transition matrix with all states and actions is not recommended. Also, our reward function is data-dependent (i.e. we use image ranking to calculate it). Q-learning \cite{sutton2011reinforcement}, which receives a state and predicts the best action, is a good fit for our task. 

Our neural network architecture is based on \cite{caicedo2015active} and is depicted in Figure \ref{fig:q_network}. 
%It receives the top result images, proxies, and history of taken actions. 
Our top images and proxies data uses the same convolution architecture composed of a convolutional layer with 8 filters of size 3x3 and a max-pooling layer. The outputs of the top images and proxies branches are concatenated with the history of actions, and projected using 3 fully connected layers to generate action scores. We employ RELU activation for the convolutional and fully connected layers. We employ convolutional layers in the top result image and proxies branches, because they capture information about image features and ordering.

Our approach also considers replay-memory to collect many data instances as it is running. Each instance is represented as [current state, new state, action, reward]. This information enriches our training data, and in each iteration, a random subset of this data is used for training. This procedure also removes short-term correlation between subsequent states, and makes our algorithm more robust and stable.

At initial stages of learning, random actions are beneficial so the agent can \emph{explore} \cite{sutton2011reinforcement} and get information about the problem. Later this information is \emph{exploited} to select actions.
%, but random actions occasionally can help to find more promising solutions.
We generate random actions with  probability decreasing from 1 to 0.1 as training progresses.

\vspace{-0.5cm}
\paragraph{Implementation.}
We implemented the described network using the Theano \cite{tool:theano}, Keras \cite{tool:keras} and DEER\footnote{https://github.com/VinF/deer/} frameworks. We use the RMSProp optimizer, a learning rate of 1e-5, and 30 epochs. At the end of each epoch, the network was evaluated on a validation set, and the network that successfully completed more searches (i.e. found the target image in at most 10 iterations) over a validation set was selected for testing.\footnote{For our Scenes dataset, the best model is acquired using percentile rank.}

% We simulate sketch drawings using edge maps \cite{Xie_2015_ICCV}. And, for user responses, we follow the same procedure from \cite{kovashka2013attribute}. We answer questions "Does the target image contains the \textit{attribute} more, equal or less than a \textit{pivot image}?" using the difference in the predicted attribute strength values for the target and pivot image. We simulate ten users adding gaussian noise with $\mu=0$ and $\sigma=1*s$, where $s$ is the standard deviation of the predicted attribute strengths.
\vspace{-0.5cm}
\section{Experimental Validation}
\label{sec:results}
\vspace{-0.2cm}

%\subsection{Experimental Setup}
%\vspace{-0.1cm}

%\vspace{-0.5cm}
\paragraph{Datasets.}
We  use  three  datasets which have frequently been used for image search:  Pubfig \cite{kumar09} with 11 attributes (e.g. smiling, rounded-face, masculine) and 769 images (after de-duplication); Scenes \cite{olvia2001, parikh2011relative} with 6 attributes (open, in perspective, etc.), and 2668 images; and Shoes \cite{kovashka2015whittlesearch} with 10 attributes (formal, high-heeled) and 12,807 images. 
We extracted fc6 deep features for Pubfig and Shoes; and fc7 features for Scenes as in \cite{modi2017confidence}.
To speed up the interaction of our reinforcement learning agent and the image retrieval system, we reduce the number of images to 1000 by clustering in the predicted attribute strengths space. 
%, which are learn by the relevance models (See section \ref{sec:setup}). 

\vspace{-0.5cm}
\paragraph{Evaluation protocol.}
For each dataset, we split the data in 70\% for training, 10\% for validation and 20\% for testing. Our reinforcement learning approach uses the train and validation splits to learn to predict actions. %The other methods are directly applied to the test set.
%as they do not require any training.
To compare the methods more precisely, we tell the user which image to search for (\textit{target image}). In each iteration, the user provides a comparison of the target and pivot/reference image, or a sketch of the target. We report percentile rank of the target, defined as the fraction of database images ranked lower than the target (in the range [0, 1], higher is better). 
%Higher percentile ranks are better, and means that the target image is near to the top ones.

\vspace{-0.5cm}
\paragraph{Baselines.}

We compare our reinforcement learning agent (RL) with three baselines:
\begin{itemize}[noitemsep,nolistsep]
    \item Whittle Search \cite{kovashka2015whittlesearch} (\textit{WS}): In each iteration, users select a (reference image, attribute) and compare target and reference for the chosen attribute dimension (``more/less/equally''). The relevance of database images which satisfy this feedback increases.
    \item Pivot round robin \cite{kovashka2013pivots} (\textit{PRR}): In contrast to \textit{WS}, \textit{PRR} provides an (image, attribute) pair, and users only need to provide a more/less/equally response.
    \item Sketch retrieval \cite{yu2016sketch} + pivot round robin \cite{kovashka2013pivots} (\textit{SK\_PRR}): In the first iteration, we ask the user for a sketch of the target image, then attribute questions follow. 
\end{itemize}

\vspace{-0.3cm}
\subsection{Simulated experiments}
\vspace{-0.1cm}

\begin{figure}[t]
    \centering
    \includegraphics[width=.32\textwidth]{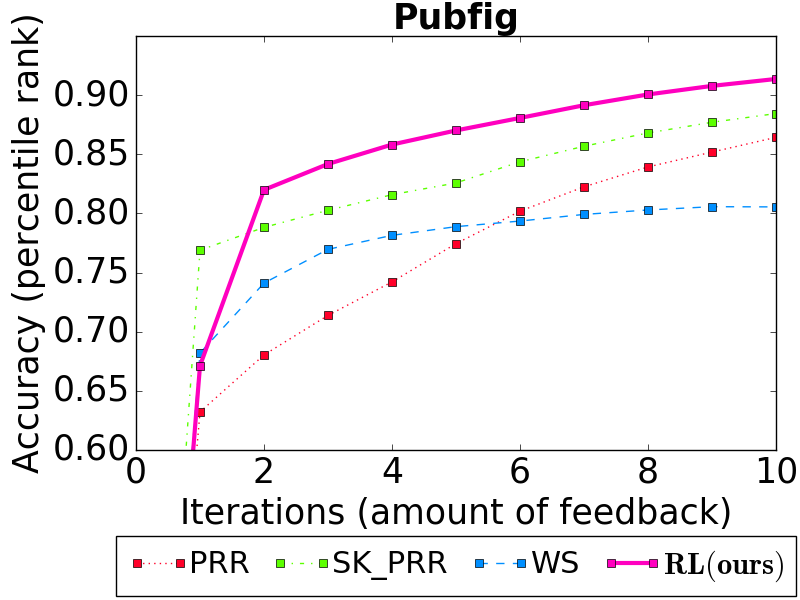}
    \includegraphics[width=.32\textwidth]{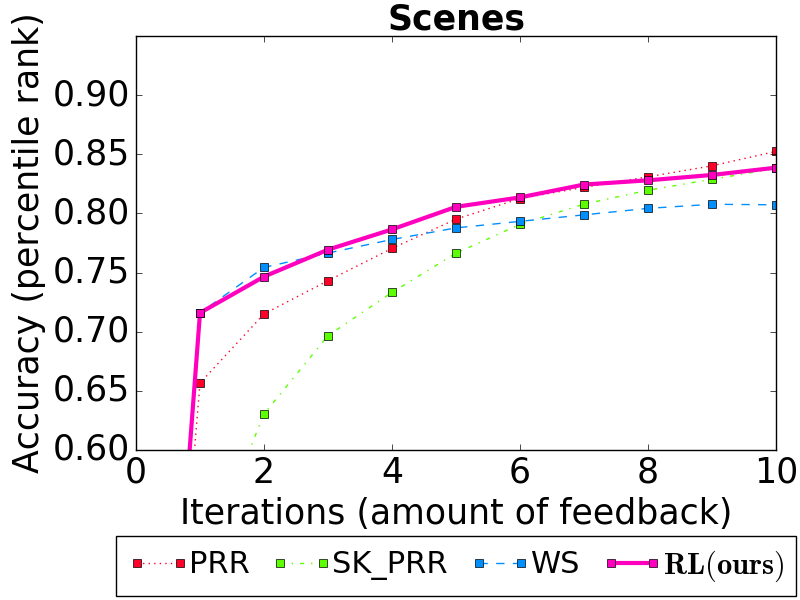}
    \includegraphics[width=.32\textwidth]{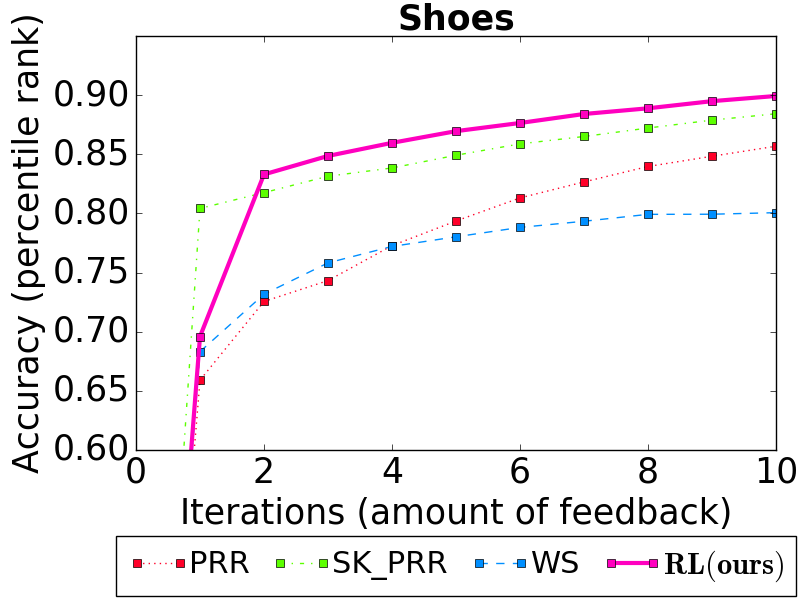}
    \vspace{-0.3cm}
    \caption{Percentile rank plots for Pubfig, Scenes, and Shoes. Our mixed-initiative RL agent outperforms the other baselines on Pubfig and Shoes, and performs competitively for Scenes. }
    %\vspace{-0.3cm}
    \label{fig:pr}
\end{figure}

\begin{table}[t]
  \centering
  \caption{AUC for percentile rank curves from Fig.~\ref{fig:pr}. Best scores are highlighted per dataset.}
  \resizebox{0.5\columnwidth}{!}
  {
    \begin{tabular}{|c|c|c|c|c|c|}
    \hline
    & PRR \cite{kovashka2013pivots}   & WS \cite{kovashka2015whittlesearch}    & SK\_PRR \cite{yu2016sketch,kovashka2013pivots} & \textbf{RL (ours)} \\
    \hline
    Pubfig & 0.729 & 0.737 & 0.789   & \textbf{0.810}     \\
    \hline
    Scenes & 0.741 & 0.741 & 0.699   & \textbf{0.754}     \\
    \hline
    Shoes  & 0.745 & 0.731 & 0.806   & \textbf{0.810}    \\
    \hline
    \hline
    avg  & 0.738 & 0.736 & 0.764   & \textbf{0.791}    \\
    \hline
    \end{tabular}
  }
  \label{tab:auc_pr}
  \vspace{-0.4cm}
\end{table}%

%In order to simulate user interactions, we require sketch drawings for \textit{SK\_PRR}, and user responses for \textit{WS} and \textit{PRR}. 
We simulate ten users as described in Sec.~\ref{sec:repr}.
Fig.~\ref{fig:pr} shows percentile rank curves for our proposed method and the three baselines. For the Pubfig and Shoes datasets, our reinforcement agent outperforms the baselines with a large margin. However, for Scenes, the improvement is reduced. Hence, we also inspect AUC for the percentile rank curves in Table \ref{tab:auc_pr}. We observe that our approach outperforms all baselines for all datasets.

% SELECTED EXPERIMENTS
% \\afs\cs.pitt.edu\projects\kovashka\nils\drive\weekly-results\2018\March\week4\results_700_1000

We observe that \textit{WS} achieves high accuracy at the very first iterations and outperforms the \textit{PRR} method. This follows the intuition that with \textit{WS}, which allows \emph{exploration}, the user can provide more meaningful feedback that reduces the search space, in contrast to earlier stages of the \textit{PRR} method.
However, in later iterations, \textit{PRR} improves accuracy because it follows a binary-search strategy iterating over all attributes. Hence, \textit{PRR} ensures diversity of feedback, in contrast to \textit{WS} which can be repetitive. \textit{SK\_PRR} outperforms \textit{WS} and \textit{PRR} in two of the three datasets. Incorporating sketch feedback enhances the informativeness of attribute-based feedback, except for Scenes. A possible explanation is that scenes are more complex than faces and shoes, as they contain more than one object. This prevents our GAN from being able to generate good photo versions of our scene edge maps (see Fig.~\ref{fig:gan}).

\vspace{-0.3cm}
\subsection{Live experiments}
\vspace{-0.1cm}
% /afs/cs.pitt.edu/usr0/nineil/private/CV/nm_caffe/search_project/shoes_live_exps/exp5

% shoes retrieval: \\afs\cs.pitt.edu\projects\kovashka\nils\drive\weekly-results\2018\Apr\week2\shoes_ret_TT

\begin{figure}[t]
    \centering
    \includegraphics[width=.32\textwidth]{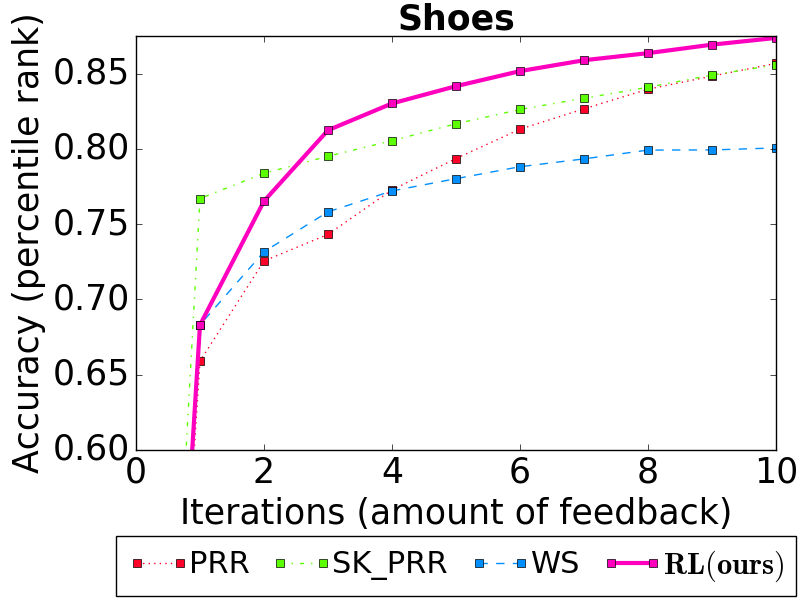}
    \includegraphics[width=.32\textwidth]{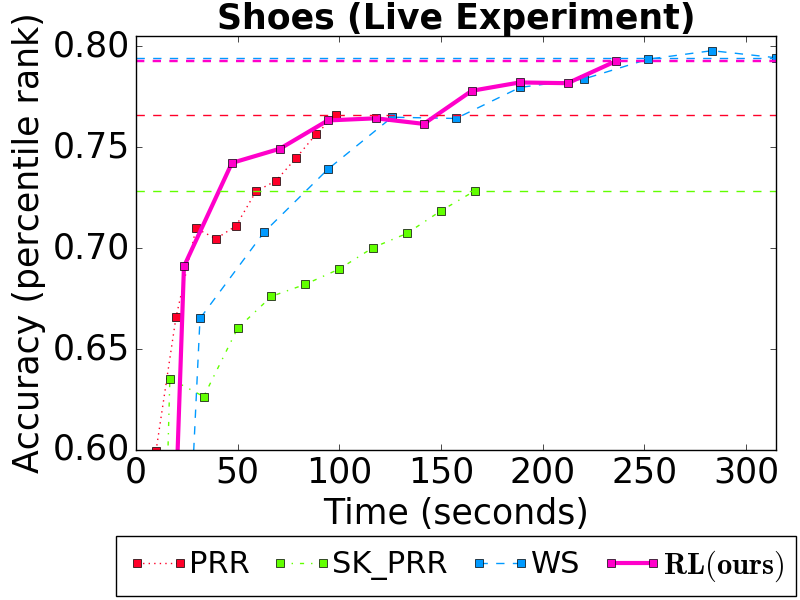}
    \vspace{-0.3cm}
    \caption{Percentile rank plots for Shoes dataset with simulated (left) and live users (right). Both experiments use sketch retrieval. Live user experiment results are plotted over time.}
    \label{fig:pr_live}
    \vspace{-0.3cm}
\end{figure}

In order to run a user study, we develop a web interface that implements our three baselines, and our approach. Our approach queries the next action using a REST API\footnote{https://blog.keras.io/building-a-simple-keras-deep-learning-rest-api.html}, that connects to our web interface. 
For this experiment, we replace sketch-to-photo coloring with sketch retrieval \cite{yu2016sketch} directly comparing features of the sketches to images, as an alternative to get diverse and realistic images. This helps avoid GPU memory problems due to multiple queries for the GAN conversion. 
%, because we use a GAN implementation in torch, and our main program is in python. 
We only conduct an experiment for the Shoes dataset because we did not find any appropriate sketch annotations for training, for Faces\footnote{Fine-grained sketches are available but most real users cannot provide such high-quality sketches.} and Scenes. The result for simulated users (Fig.~\ref{fig:pr_live} left) in this setting is similar to our previous findings: our approach outperforms all baselines.

We recruit workers on Amazon Mechanical Turk and university students to search for 100 images. Each participant searches for one image, which is the same for the four methods. We request Turkers with location in the US, HIT approval rate greater or equal to 98\%, and at least 1000 approved HITs. We remove blank and careless sketches (i.e. just straight lines), which results in 88 searches.
% We conduct a user study with 100 images using Amazon Mechanical Turk, . Each searches for five images which are the same over the four methods. 
% We conduct a user study with seven participants. Each searches for five images which are the same over the four methods. 
The results are shown in Fig.~\ref{fig:pr_live} (right). Because different interactions require very different amount of user time ($PRR$: 9s, $SK\_PRR$: 16s, $WS$: 31s, and $RL$: 23s), we plot time on the x-axis, multiplying each iteration by the number of seconds it requires. 
We show horizontal lines with the final (highest) percentile rank a method achieves. Our $RL$ method and $WS$ achieve similar peak performance (79.2\% for $RL$ and 79.4\% for $WS$) while $PRR$ only achieves 76.6\% at the end of 10 iterations. However, our method achieves higher performance early on; the curve for $RL$ is higher than that for $WS$ until about 230s of user time spent, then performance is similar.
%We observe that our approach $RL$ outperforms $PRR$ and $SK\_PRR$. Only $WS$ has a slightly similar performance to our approach, however, the users need more time to provide more informative feedback. $WS$ requires $1.6$ times the time of our approach. 
Thus, our approach achieves higher performance in a smaller amount of time, compared to the strongest baseline $WS$. 

We include sketches provided from our live users in our supp. material.

\vspace{-0.3cm}
\subsection{Qualitative Results}
\vspace{-0.1cm}

\begin{figure}[t]
    \centering
    % pubfig
    \includegraphics[width=.09\textwidth]{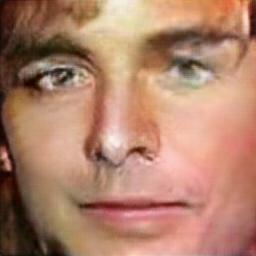}
    \includegraphics[width=.09\textwidth]{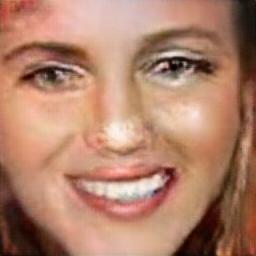}
    \includegraphics[width=.09\textwidth]{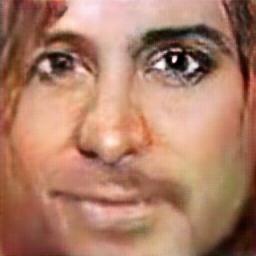}
    %shoes
    \includegraphics[width=.09\textwidth]{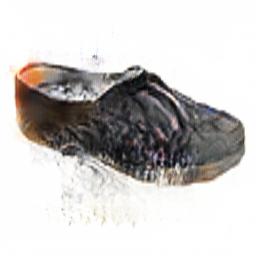}
    \includegraphics[width=.09\textwidth]{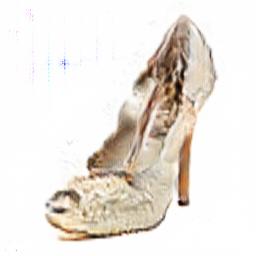}
    \includegraphics[width=.09\textwidth]{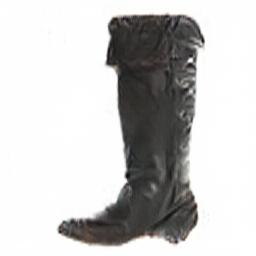}
    % osr
    \includegraphics[width=.09\textwidth]{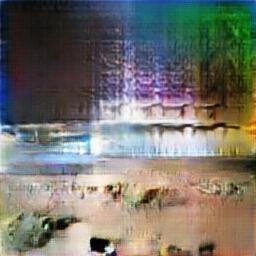}
    \includegraphics[width=.09\textwidth]{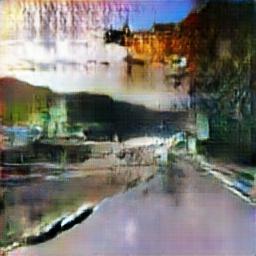}
    \includegraphics[width=.09\textwidth]{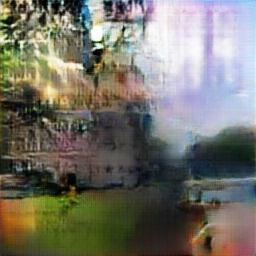}\\
    \vspace{-0.3cm}
    \caption{Sample sketch-to-photo colored images for Pubfig (columns 1-3), Shoes (columns 4-6), and Scenes (columns 7-9). Each column denotes a different category.}
    %\vspace{-0.6cm}
    \label{fig:gan}
\end{figure}

\begin{figure}[t]
    \centering
    \includegraphics[width=.32\textwidth]{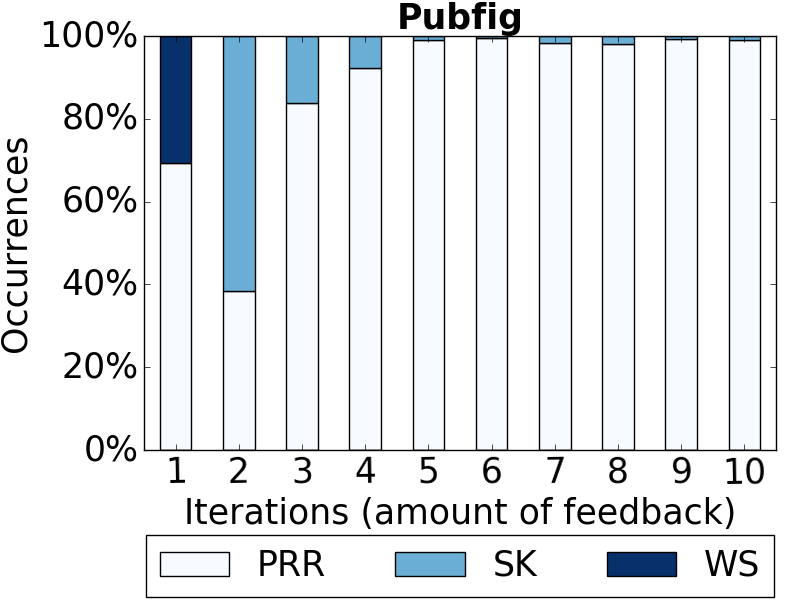}
    \includegraphics[width=.32\textwidth]{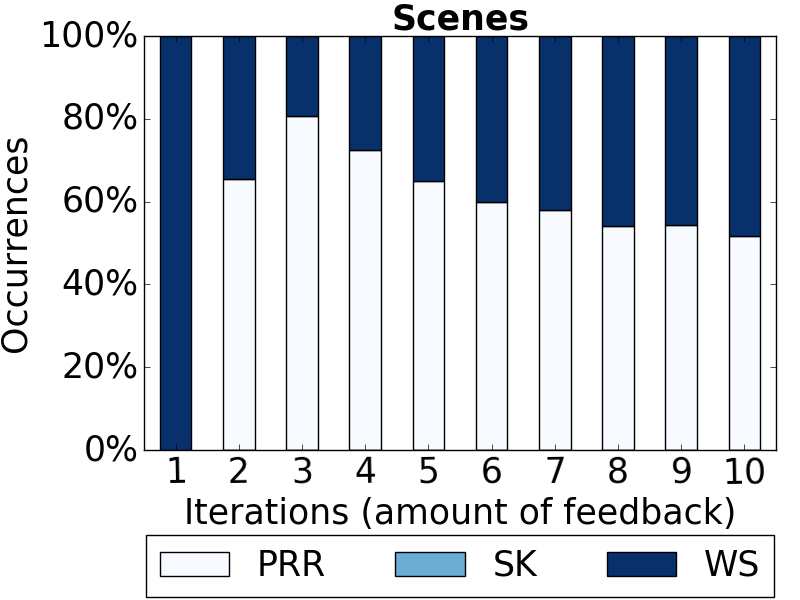}
    \includegraphics[width=.32\textwidth]{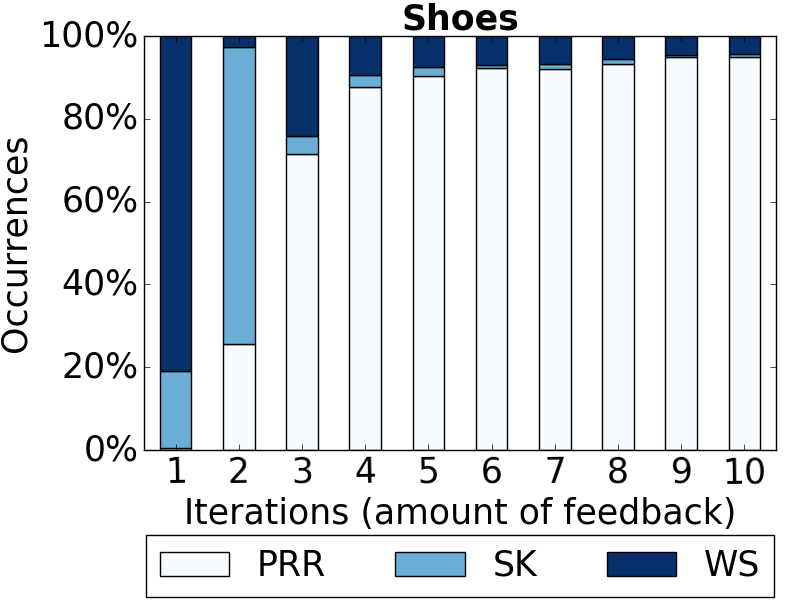}
    \vspace{-0.3cm}
    \caption{Percentage of actions predicted by our approach in the test set.}
    \vspace{-0.4cm}
    \label{fig:rl_acts}
\end{figure}

In order to understand the success of our approach, we visualize some of the generated colored pictures (Fig.~\ref{fig:gan}), and we also show the predicted actions on our test split (Fig.~\ref{fig:rl_acts}).

For our sketch-to-photo generated images, we observe that the most realistic ones correspond to Pubfig, then Shoes, and finally Scenes. This order also corresponds with the performance of our method in terms of percentile rank, where Pubfig and Shoes achieve the best performance. Scenes did not benefit from the generated images as much because they are not realistic and present poor quality. However, our GAN associates brown color to coast (Fig.~\ref{fig:gan},~column 7). Similarly, it learns green color for forest (Fig.~\ref{fig:gan},~column 9). We add more visualizations in our supp. material including edge maps.% and sketches.

We also want to understand our mixed-initiative RL agent, so we count its predicted actions per iteration in Fig.~\ref{fig:rl_acts}. 
%, in addition to the occurrences when an image is found. In all charts, we observe that percentage of found images increases as the amount of feedback (See yellow bars). This is a sign that our agent learns to prioritize meaningful feedback to find more images. 
For Pubfig and Shoes, we observe that \emph{SK} (sketch) and \textit{WS} actions are mainly performed in iterations 1 and 2, because these are the \emph{exploration}-like actions. Then, after iteration 3, the \textit{PRR} is the most common one. Once the most beneficial human knowledge is acquired, having a computer suggest feedback (in the form of questions) helps reduce the search space the fastest. Hence, our agent learned to prioritize human-initiated feedback early on, and complement it with machine-initiated feedback in later iterations.
For Scenes, our method %discards the sketch action, 
prioritizes $WS$ early on and $PRR$ later, and ignores $SK$ because it does not provide much benefit.
%and performs \textit{WS} actions for the first iteration, then \textit{WS} and \textit{PRR} actions are performed in later iterations. It discards sketch actions, 
%, as shown in Figs.~\ref{fig:pr} and \ref{fig:gan}.

\vspace{-0.3cm}
\section{Conclusion}
\label{sec:conclusion}
\vspace{-0.1cm}

We explored the problem of selecting interactions in a mixed-initiative image retrieval system. Our approach selects the most appropriate interaction per iteration using reinforcement learning. We find that our model prefers human-initiated feedback in former iterations, and complements it with machine-based feedback requests (e.g. questions) in later iterations. We outperform standard image retrieval approaches with simulated and real users.
For future work, we plan to learn personalized reinforcement agents that follow the individual attribute interpretations \cite{Kovashka15a, murrugarra2017}, visual perception and sketching style of users.

\bibliography{egbib}
\end{document}